\pdfoutput=1

\documentclass[11pt]{article}

\usepackage[]{acl}

\usepackage{times}
\usepackage{latexsym}

\usepackage[T2A,T1]{fontenc}
\usepackage[utf8]{inputenc}
\usepackage[russian,finnish,english]{babel}

\usepackage{microtype}

\usepackage{inconsolata}

\usepackage{graphicx}

\title{DAG: Dictionary-Augmented Generation for Disambiguation of Sentences in Endangered Uralic Languages using ChatGPT}

\author{Mika Hämäläinen \\
  Metropolia University of Applied Sciences \\
  Helsinki, Finland \\
  \texttt{first.last@metropolia.fi} \\}

\begin{document}
\maketitle
\begin{abstract}
We showcase that ChatGPT can be used to disambiguate lemmas in two endangered languages ChatGPT is not proficient in, namely Erzya and Skolt Sami. We augment our prompt by providing dictionary translations of the candidate lemmas to a majority language - Finnish in our case. This dictionary augmented generation approach results in 50\% accuracy for Skolt Sami and 41\% accuracy for Erzya. On a closer inspection, many of the error types were of the kind even an untrained human annotator would make.
\end{abstract}

\section{Introduction}

Morphological disambiguation is a critical task in natural language processing (NLP), especially for morphologically rich and endangered languages. Skolt Sami (sms) and Erzya (myv), both belong to Uralic language family and they are classified as critically and definitely endangered respectively by Unesco \cite{moseley_2010}. This poses significant challenges in this domain due to their complex morphological systems and limited available linguistic resources (see \citealt{hamalainen2021endangered}). In languages like these, each word form can have multiple possible morphological interpretations and lemmas, and determining the correct one in context is essential for accurate language processing.

Traditional approaches to morphological disambiguation for Uralic languages often rely on finite-state transducers (FSTs) and constraint grammars (CGs) that list all potential lemmas for a word, but these systems struggle to accurately select the appropriate lemma in ambiguous contexts - not to mention that CG disambiguators have not even been created to a majority of these languages. Additionally, while some modern NLP techniques, such as machine learning models, have been successful in languages with large datasets (see \citealt{shen-etal-2016-role,zalmout2017don}), such methods are less effective for languages like Skolt Sami and Erzya, which suffer from limited annotated corpora and lexicographical resources.

This paper presents a novel method for performing morphological disambiguation for Skolt Sami and Erzya that leverages a combination of a traditional FST-based analyzer, a bilingual dictionary and a state-of-the-art language model, namely ChatGPT. Our approach involves passing each sentence through an FST to generate a list of possible lemmas for every word. These lemmas are then translated into a majority language (Finnish in our case) using a dictionary.  Finally, we utilize ChatGPT, a powerful transformer-based language model, to analyze the translated sentence, disambiguate the lemmas, and select the most contextually appropriate form for each word. The dictionary needs to be provided given that ChatGPT is not proficient in these languages.

By integrating the structured linguistic knowledge from FSTs with the contextual understanding of large language models, this method aims find a novel way that does not need a time consuming rule-writing or data annotation process for morphological disambiguation for Skolt Sami and Erzya. The proposed approach is particularly valuable for endangered languages, where data scarcity hinders the development of purely data-driven models. This paper details the methodology, presents an evaluation of the approach, and discusses the potential for applying this approach to other morphologically complex languages. We have released our disambiguation code as an addition to UralicNLP\footnote{https://github.com/mikahama/uralicNLP/wiki/Disambiguation}.

\begin{table*}[!ht]
\centering
\resizebox{\textwidth}{!}{%
\begin{tabular}{|l|l|}
\hline
Prompt template                                                                                                                                                                                                                                                                                                                                                                                                                                                                                                                                                                                                                                                                                                                                                                                                                                              & Actual prompt                                                                                                                                                                                                                                                                                                                                                                                                                                                                                                                                                                                                                                                                                                                                                                                                                                                                                                                                                                                                                                                                                                                                                                                                                                                                                                                                                                                                                                                                                                                                                                                                                                                                                                                                          \\ \hline
\begin{tabular}[c]{@{}l@{}}Your task is to disambiguate a sentence in {[}LANGUAGE{]} You will be given the sentence,\\ a table that has all of the words of the sentence in separate rows and a comma separated\\ list of possible lemmas. You will need to pick the correct lemma for each word so that\\ every word will have only one lemma. To help you understand {[}LANGUAGE{]} you will\\ also get a second table that gives you translations of the words in {[}LANGUAGE2{]}.\\ \\ Sentence:\\ {[}SENTENCE{]}\\ \\ Table of lemmas:\\ \\ {[}TABLE1{]}\\ \\ \\ {[}LANGUAGE{]} - {[}LANGUAGE2{]} vocabulary:\\ \\ {[}TABLE2{]}\\ \\ \\ Please write out the steps of your decision process and provide a list of lemmas \\ in JSON format at the very end of your answer. \\ Example: \{"lemmas": {[}"lemma 1", "lemma 2", "lemma 3"{]}\}\end{tabular} & \begin{tabular}[c]{@{}l@{}}Your task is to disambiguate a sentence in Skolt Sami. You will be given the sentence,\\ a table that has all of the words of the sentence in separate rows and a comma separated\\ list of possible lemmas. You will need to pick the correct lemma for each word so that\\ every word will have only one lemma. To help you understand Skolt Sami you will\\ also get a second table that gives you translations of the words in Finnish.\\ \\ Sentence:\\ Päärna mõ\textsuperscript{$\prime$}nne mååusat .\\ \\ Table of lemmas:\\ \\ +--------------------------+\\ |  Word |      Lemmas      |\\ +-------+------------------+\\ | Päärna|      päärnaž     |\\ +-------+------------------+\\ | mõ\textsuperscript{$\prime$}nne|mõõnnâd, mõ\textsuperscript{$\prime$}nn'jed|\\ +-------+------------------+\\ |mååusat|      mååusat     |\\ +-------+------------------+\\ |   .   |         .        |\\ +--------------------------+\\ \\ Skolt Sami - Finnish vocabulary:\\ \\ +---------------------------------------+\\ |Skolt Sami|           Finnish          |\\ +----------+----------------------------+\\ |  päärnaž |poikanen, lapsi, pieni poika|\\ +----------+----------------------------+\\ |  mõõnnâd |            mennä           |\\ +----------+----------------------------+\\ | mõ\textsuperscript{$\prime$}nn'jed|           munata           |\\ +----------+----------------------------+\\ |  mååusat |          takaisin          |\\ +----------+----------------------------+\\ |     .    |                            |\\ +---------------------------------------+\\ \\ Please write out the steps of your decision process and provide a list of lemmas\\ in JSON format at the very end of your answer. \\ Example: \{"lemmas": {[}"lemma 1", "lemma 2", "lemma 3"{]}\}\end{tabular} \\ \hline
\end{tabular}%
}
\caption{The prompt template and an example of it filled}
\label{tab:prompt-template}
\end{table*}

\section{Related Work}

Constraint grammars (CGs) \cite{karlsson1990constraint} have been widely used in the context of Uralic languages for disambiguation given their compatibility with the output of FSTs. They are, however, not used widely anymore in the mainstream NLP research. In this section, we will go through some of the more modern NLP approaches to this task in the context of endangered languages.

In a work focusing on Uralic languages \cite{ens2019morphosyntactic}, the authors propose a Long Short-Term Memory (LSTM) model that automatically ranks morphological readings of sentences based on their quality. This ranking can be used either to evaluate existing CG disambiguators or to directly disambiguate sentences. Notably, their approach relies on morphological abstraction and can be effectively trained with minimal data.

Apertium's approach \cite{khanna2021recent} is to employ statistical methods based on patterns learned from a corpus in addition to CG. Two prominent methods include a bigram-based first-order Hidden Markov Model (HMM), which selects analyses based on a probabilistic model of part-of-speech tag sequences in context, and an Averaged Perceptron tagger, which assigns weights to features defined by language-pair developers.

The paper by \citet{keleg-etal-2020-unsupervised} introduces a method for weighting the outputs of an FST-based morphological analyzer to disambiguate its results. The approach uses a word2vec model, trained in an unsupervised manner on raw, untagged corpora, to capture semantic meaning. Unlike traditional methods that require manually constructed tagged corpora, this method disambiguates morphological analyses without relying on such resources. Additionally, it focuses on token-level information rather than context, differing from most approaches that heavily depend on contextual features for disambiguation.

\section{Method}

The method itself does not require any training or additional annotated data. However, to evaluate our method, we use the Universal Dependences treebanks for Erzya \cite{rueter2018towards} and Skolt Sami \cite{nivre2022ud_skolt_sami}. These treebanks have word forms and their correct lemmas for each word in each sentence. Given that we do not need to do training, we concatenate the training and test datasets into one dataset for both languages.

Next, we used the Skolt Sami \cite{rueter2020fst} and Erzya \cite{rueter2020open} FSTs through PYHFST \cite{alnajjar2023pyhfst} to get all possible lemmas for each word form in the treebanks. For every lemma, we look up its Finnish translations from the Akusanat dictionary \cite{hamalainenadvances}.

As we build our corpus of sentences to be disambiguated this way, we need to do some filtering. If a sentence does not have any ambiguity, all of it's words lemmatized by the FST or all of the potential lemmas mapped to at least one Finnish translation, the sentence is removed from the corpus. This way, we end up with 40 sentences from Skolt Sami treebank and 17 sentences form Erzya treebank.

The reason why this filtering is done is that ChatGPT has no proficiency in Erzya or Skolt Sami. It does not make sense to try this disambiguation if we cannot provide ChatGPT with enough information so that we can even assume that it would be capable of disambiguating the sentence its given.

Our corpus is used to populate a carefully planned prompt template as seen in Table \ref{tab:prompt-template}. The template consists of 5 parts. The first part is the task description where the LLM is given the task instruction. The second part has the sentence that needs to be disambiguated. The third part is a table that maps word forms to their possible lemmas. The fourth part is a table that maps lemmas to their Finnish translations. The fifth part instructs the model to take time in making its decision and to return the final result in a certain JSON format.

The tables are formatted in Markdown format. We use OpenAI API and select gpt-4o as the model to be used with temperature set to the default of 1. Every sentence is prompted separately so that the results won't have an influence on each other. This experiment cost us \$0.41. 

\section{Results}

If we calculate the accuracy of the disambiguation by ChatGPT on a sentence level, meaning the number of fully correctly disambiguated sentences out of all the sentences, we get the following results: \textbf{Skolt Sami 50\% and Erzya 41\% accuracy}. At this step, it is already worth noting that ChatGPT omitted some punctuations in its lemmatization. Also, Skolt Sami has several unique Unicode characters that look identical to other Unicode characters. ChatGPT had a tendency of sometimes changing the original characters to their lookalikes in the output. These cases were still counted as correct answers.

Most of the erroneous sentences had only one word that exhibited ambiguity. In the case of several ambiguous words, ChatGPT almost always made a mistake for only one word in the sentence. There were, however, some cases where the same word was repeated twice, in which case ChatGPT lemmatized both of them wrong following the same logic.

\subsection{Error analysis}

In this section, we take a closer look at the errors ChatGPT made when disambiguating between lemmas in Skolt Sami and Erzya.

\subsubsection{Derivational forms}

Derivational forms caused problems in Erzya but not in Skolt Sami. All in all, there were 5 of these cases. Here is an example of an erroneous attempt of picking the correct lemma by ChatGPT:

\begin{table}[ht]
\centering
\begin{tabular}{l}
\begin{tabular}[c]{@{}l@{}}6. \foreignlanguage{russian}{**Омбоцеде**}\\    - Lemma options: \foreignlanguage{russian}{омбоцеде, омбоце}\\    - Translations: "toista kertaa" (Finnish for "second time")\\ and "toinen" (Finnish for "second/another")\\ - In this sentence, \foreignlanguage{russian}{"омбоцеде"} seems to mean \\ "toista kertaa," referring to the concept of "second time" \\ rather than just "another."\\    - Therefore, we'll choose \foreignlanguage{russian}{"омбоцеде"} as the lemma.\end{tabular}
\end{tabular}
\end{table}

In the example, \foreignlanguage{russian}{омбоцеде} can either be interpreted as an adverb and thus a lemma on its own right or as a derivational form of the numeral \foreignlanguage{russian}{омбоце}.

\subsubsection{Near synonyms}

Words that almost mean the same thing were also a problem source. There were 2 of these cases for Erzya and 4 for Skolt Sami. Here is an example from ChatGPT's output for Skolt Sami:

\begin{table}[ht]
\centering
\begin{tabular}{l}
\begin{tabular}[c]{@{}l@{}}12. **ǩeäčč**:\\     - "ǩiččâd" means "katsoa" (to look) while "ǩiõččâd"\\  translates similarly, also as "to browse".\\     - Both could make sense but as an indicative action \\ following conjunction, "ǩiččâd" aligns well.\\     - **Chosen Lemma**: "ǩiččâd"\end{tabular}
\end{tabular}
\end{table}

\subsubsection{Lack of context}

Sometimes the sentence itself was not quite enough to disambiguate the correct lemma as both lemma candidates remained viable. There were 6 of these cases in Skolt Sami and 2 in Erzya. Below is an Erzya example of such a case:

\begin{table}[ht]
\centering
\begin{tabular}{l}
\begin{tabular}[c]{@{}l@{}}- \foreignlanguage{russian}{**Арасть:**}\\    - Here we have a choice between \foreignlanguage{russian}{"арамс"} \\(to become) and \foreignlanguage{russian}{"арась"} (no/missing).      \\ - Contextual understanding is needed.\\       - With no other context suggesting negation or \\ anything missing, it's more plausible \foreignlanguage{russian}{"арасть"} \\ relates to "arams" (to become) especially if the  \\ sentence forms a complete statement.\end{tabular}
\end{tabular}
\end{table}

\subsubsection{Failure to transfer POS information}

This issue did not happen in Erzya, but it did happen in Skolt Sami 3 times. In these cases, the correct answer was rather clear based on the parts-of-speech of the Finnish words. However, ChatGPT seemed not to take this piece of information in consideration. Here is an example:

\begin{table}[ht]
\centering
\begin{tabular}{l}
\begin{tabular}[c]{@{}l@{}}5. **puälddmõõžž**\\    - We have two possible lemmas: "puä\textsuperscript{$\prime$}ldded" \\ (paahtaa, polttaa) meaning "to roast, burn" \\ and "puälddmõš" (polttaminen) meaning \\ "burning." The sentence seems to involve \\ actions, so "puä\textsuperscript{$\prime$}ldded," which denotes an \\ action, fits better in this context.\\    - **Chosen lemma:** puä\textsuperscript{$\prime$}ldded\end{tabular}
\end{tabular}
\end{table}

In the above example, puä\textsuperscript{$\prime$}ldded is a verb and puälddmõš is a noun. It is possible that the fact that ChatGPT translates the latter into an English noun that is derived from a verb makes ChatGPT forget about the parts-of-speech the Finnish translations reveal.

\subsubsection{FST issues}

There was 1 case in Skolt Sami where the correct lemma was not among the lemmas produced by the FST. This gave ChatGPT no chance of getting the lemma right. Take a look at this example:

\begin{table}[ht]
\centering
\begin{tabular}{l}
\begin{tabular}[c]{@{}l@{}}5. **Ǩeârdd**:\\    - Possible lemma: `ǩiõrddâd`\\    - Finnish translation: "sietää, kärsiä, \\ kestää" (endure, suffer, withstand)\\    - Analysis: There is only one lemma \\ provided. It suggests an action or event \\ being endured.\\    - Conclusion: The lemma for \\ "ǩeârdd" is "ǩiõrddâd".\end{tabular}
\end{tabular}
\end{table}

The correct lemma would have been ǩe\textsuperscript{$\prime$}rdded, which was not even an option.

\subsubsection{Lack of morphosyntactic understanding}

There were 5 of this type of errors in Skolt Sami and 1 in Erzya. In these cases, ChatGPT could have picked the correct answer had it known more about morphology and morphosyntax. Below is an example of this case:

\begin{table}[ht]
\centering
\begin{tabular}{l}
\begin{tabular}[c]{@{}l@{}}**First word: "Tiâr"**\\      - Potential lemmas: "tiârr" and "tiârrâd"\\       - Translation in Finnish: "tiârr" means \\  "laukka" (noun, meaning "gallop").\\      - "tiârrâd" means "laukkoa, juosta, laukata"\\  (verb forms, meaning "to gallop, to run, \\ to canter").\\      - The context "tiâr" is paired with "olgglab", \\suggesting  it is more likely to be a noun as\\ "to gallop farther"  makes less sense\\ without additional  context.\\      - Choosing lemma: "tiârr"  (noun - gallop)\end{tabular}
\end{tabular}
\end{table}

\subsubsection{Hallucination}

In one case, ChatGPT hallucinated a lemma for Skolt Sami that was not one of the options. In this example, ChatGPT forgot the provided lemma päärnaž.

\begin{table}[ht]
\centering
\begin{tabular}{l}
\begin{tabular}[c]{@{}l@{}}1. **Päärna**:\\    - Lemmas: "pärarnaž"\\    - Finnish translation: "poikanen, lapsi, pieni poika"\\    - "Päärna" directly corresponds to "pärarnaž," which \\ means "a small boy" or "child" in Finnish. There is only \\ one lemma option, so it is straightforward.\end{tabular}
\end{tabular}
\end{table}

\section{Conclusions}

We have demonstrated a viable method for conducting disambiguation on endangered language data. The results are very good given that ChatGPT is not proficient in Skolt Sami or Erzya. If we look at the errors, they mostly make sense to a human. Many of the error types are such that even a novice human annotator without training in these languages would make similar mistakes.

It is important that we, in the endangered NLP community, keep our eyes and minds open, and embrace the new potential in LLMs. Perhaps they don't speak our languages of interest yet, but they can still make reasoned decisions if enough information is provided to them.

\bibliography{acl_latex}

\end{document}